\documentclass[letterpaper,10pt,conference]{ieeeconf}

\usepackage[utf8]{inputenc}
\let\labelindent\relax
\usepackage{times} \usepackage{graphicx} \usepackage{amsmath,amssymb} \usepackage{booktabs} \usepackage{enumitem} \usepackage{cite} \usepackage{url} \usepackage{algorithm} \usepackage{algorithmic} \usepackage{adjustbox} \usepackage{comment} \usepackage{multirow} \usepackage{caption} \usepackage{placeins}

\setlist[itemize]{noitemsep, topsep=2pt, leftmargin=*}

\title{Write-Safe Flow Field Mapping under Ambiguous Onboard Sensing and Localization Drift}

\author{
\authorblockN{Linhao Jin\authorrefmark{1},
Qimin Feng\authorrefmark{1},
Peter Gunnarson\authorrefmark{2},
Qiang Zhong\authorrefmark{1}}
\authorblockA{\authorrefmark{1}Mechanical Engineering Department, Iowa State University, Ames, IA, USA}
\authorblockA{\authorrefmark{2}School of Engineering, Brown University, Providence, RI, USA}
\authorblockA{Email: linhao@iastate.edu, qmfeng11@iastate.edu, peter\_gunnarson@brown.edu, qzhong1@iastate.edu}
}

\begin{document} \maketitle \begin{abstract} Mobile robots can infer local flow structure from onboard sensing, but a locally plausible estimate is not always safe to write into a global map. Similar flow structures may produce ambiguous observations, while localization drift causes predicted patches to be written at incorrect locations. Repeated misregistered updates then accumulate into persistent ghost structures.

We address this failure mode with a map-reference-aware conservative fusion framework. The model predicts a local velocity patch and a learned write-safety score that continuously attenuates uncertain map updates while permitting initialization when no reliable map reference is available.

Across synthetic jet and crossflow environments, the proposed method reduces average ghost contamination by 42\% relative to ungated fusion. A zero-shot hardware replay using real pressure and optical-flow measurements from a thruster wake further reduces ghost contamination by 39\% while retaining 81\% map coverage. These results show that safe map writing is critical for flow mapping under ambiguous sensing and localization drift. \end{abstract}

\begin{figure}[!t] \vspace{-0.5em} \centering \includegraphics[width=0.95\columnwidth]{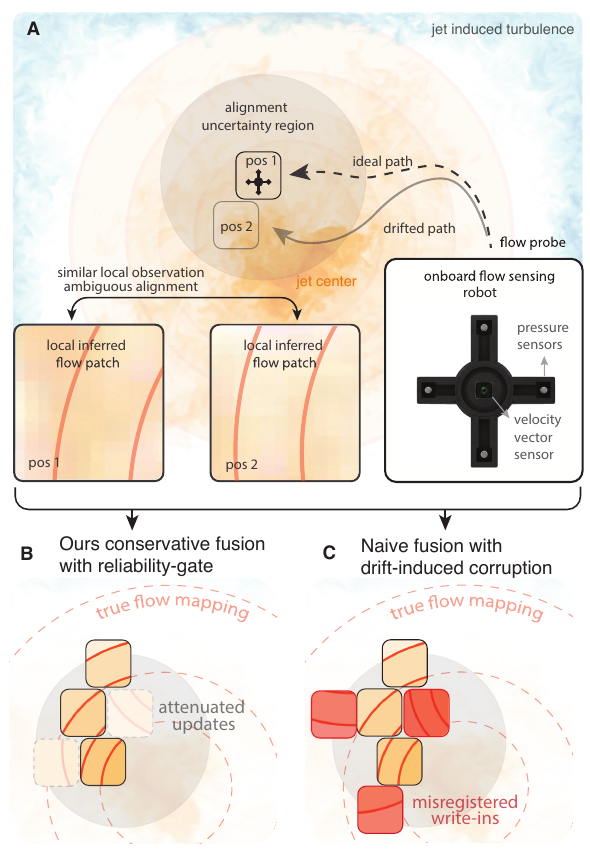} \caption{ Drift-induced map corruption under aliased flow sensing. \textbf{(A)} Similar local flow observations create ambiguous local-to-global alignment, while localization drift shifts their reported positions. \textbf{(B)} Reliability-gated conservative fusion attenuates uncertain writes to preserve global map consistency. \textbf{(C)} Ungated fusion accumulates misregistered patches into persistent ghost structures.} \label{fig:overview} \vspace{-1.5em} \end{figure}


\section{Introduction}

Flow is more than a disturbance to a mobile robot. In oceans, rivers, and other flow-dominant environments, structured currents influence vehicle motion and sampling strategies~\cite{petres2007path}, while mapping and measuring these environments remain central to ocean observation~\cite{wolfl2019seafloor,germanovich2015measuring}. Robots that interpret hydrodynamic cues can also exploit ambient flow for navigation~\cite{gunnarson2021learning}. Yet most robotic systems still compensate for flow rather than map it.

The main difficulty is not simply sparse sensing. Local flow observations are often not unique. Near-body pressure and velocity measurements describe only a small neighborhood around the robot (Fig.~\ref{fig:overview}A). Biological systems can extract informative hydrodynamic cues from such measurements~\cite{pohlmann2001tracking,dehnhardt2001hydrodynamic,liao2007review}. Artificial lateral-line systems have likewise demonstrated hydrodynamic imaging and source localization from local flow measurements~\cite{yang2006distant,wolf2019hydrodynamic}. However, sparse local measurements can be spatially non-unique: coherent, symmetric, or repeated flow structures may generate similar sensor signatures at different locations. We refer to this spatial non-uniqueness as \emph{observation aliasing}. Under aliasing, a locally plausible flow estimate does not uniquely identify where it belongs.

Localization drift turns this ambiguity into map corruption. Underwater robots generally lack continuous satellite-based positioning and therefore rely on dead reckoning, acoustic positioning, or other signal- and map-based navigation methods~\cite{zhang2023auvnavigation}. A predicted flow patch may therefore be locally correct but written at the wrong global location. Repeated errors accumulate into coherent but nonexistent ``ghost'' structures (Fig.~\ref{fig:overview}C). Unlike transient prediction noise, a misregistered write persists. Worse, the corrupted map becomes the reference for later observations and can trigger further incorrect updates.

Existing methods address flow inference or localization, but not the safety of recursive map updates. Folk \emph{et al.} infer local wind fields from onboard sensing~\cite{folk2024learning}, while Kumar \emph{et al.} jointly estimate robot trajectories and parametric marine flow models~\cite{kumar2025flow}. Probabilistic SLAM manages pose uncertainty using geometric constraints and global optimization~\cite{thrun2005probabilistic,cadena2016past,bailey2006consistency}, but persistent references may be sparse in open water. Learned and physics-informed flow models provide expressive field representations~\cite{pfaff2020learning,kochkov2021machine,raissi2019physics}, but typically target offline reconstruction or local prediction. These approaches do not directly determine whether an incoming flow estimate is safe to commit to an evolving map under ambiguous alignment.

We therefore pose online flow mapping as a write-safety problem. The robot must decide not only what to predict, but whether the prediction is spatially safe to write. We introduce a map-reference-aware conservative fusion framework (Fig.~\ref{fig:overview}B). The model predicts a local velocity patch, its informativeness, and a write-safety score $\kappa_t$. A continuous gate attenuates uncertain updates when a valid map reference exists, while still permitting map initialization when evidence is unavailable. The objective is not to fill every grid cell, but to preserve a spatially consistent map under drift.

Our contributions are as follows:
\begin{itemize}
\item We introduce a map-reference-aware continuous fusion rule that separates patch informativeness from write safety.
\item We validate the framework across synthetic flow scenes and in zero-shot hardware replay using real pressure and optical-flow measurements.
\end{itemize}

\section{Methods} \label{sec:methods}

Our framework maintains an evolving flow map, predicts a local velocity patch and its utility, and attenuates updates that cannot be reliably aligned with the map.

\subsection{Problem Formulation and Map Counterpart} \label{sec:problem}

We map a horizontal domain $\mathcal{D}\subset\mathbb{R}^2$ from onboard flow sensing. The true sensor pose is $\mathbf{p}_t^\star$, while odometry reports $\hat{\mathbf{p}}_t=\mathbf{p}_t^\star+\boldsymbol{\eta}_t$. Fusing patches at the drifting pose $\hat{\mathbf{p}}_t$ can therefore produce spatially misregistered ghost structures.

The map state is $\mathcal{M}_t=(\boldsymbol{\Omega}_t,\Psi_t)$, where $\boldsymbol{\Omega}_t:\mathcal{D}\rightarrow\mathbb{R}^2$ stores velocity and $\Psi_t:\mathcal{D}\rightarrow[0,1]$ stores accumulated evidence. The onboard observation contains four pressure measurements and a $3\times3$ optical-flow velocity stencil: \begin{equation} \mathbf{o}_t= (\boldsymbol{\pi}_t,\mathbf{U}_t^{\mathrm{of}}), \qquad \boldsymbol{\pi}_t\in\mathbb{R}^4,\quad \mathbf{U}_t^{\mathrm{of}}\in\mathbb{R}^{2\times3\times3}. \label{eq:observation} \end{equation}

Because the map stores velocity, its local counterpart is defined only for the velocity stencil. For stencil footprint $\mathcal{F}_t=\mathcal{F}(\hat{\mathbf{p}}_t)$, map-reference confidence and the corresponding stencil are \begin{align} c_t^{\mathrm{map}} &= \frac{1}{|\mathcal{F}_t|} \sum_{(x,y)\in\mathcal{F}_t} \Psi_{t-1}(x,y), \label{eq:cmap}\\ \tilde{\mathbf{U}}_t^{\mathrm{map}} &= \begin{cases} \mathcal{Q}_v (\boldsymbol{\Omega}_{t-1},\hat{\mathbf{p}}_t), & c_t^{\mathrm{map}}>\tau_{\Psi},\\ \mathbf{U}_{\emptyset}, & c_t^{\mathrm{map}}\leq\tau_{\Psi}, \end{cases} \label{eq:utilde_deploy} \end{align} where $\mathcal{Q}_v$ samples a $2\times3\times3$ stencil and $\mathbf{U}_{\emptyset}$ is a learned null token. Thus, missing map evidence is represented as an unavailable reference rather than as disagreement.

\subsection{Patch Prediction and Conservative Fusion} \label{sec:prediction_fusion}
Pressure is encoded by $\phi_\pi$, while onboard and map velocity stencils share $\phi_v$. Both are three-layer MLPs
($D_{\mathrm{in}}\!\rightarrow64\!\rightarrow64\!\rightarrow32$). Their encoded features, confidence-weighted discrepancy, and recurrent input are
\begin{align}
\mathbf{z}_t^{\pi}
&=\phi_{\pi}(\boldsymbol{\pi}_t),
\quad
\mathbf{z}_t^v
=\phi_v(\mathbf{U}_t^{\mathrm{of}}),
\quad
\tilde{\mathbf{z}}_t^v
=\phi_v(\tilde{\mathbf{U}}_t^{\mathrm{map}}),
\label{eq:encoders}\\
\mathbf{d}_t
&=c_t^{\mathrm{map}}
\left|\mathbf{z}_t^v-\tilde{\mathbf{z}}_t^v\right|,
\label{eq:confidence_difference}\\
\mathbf{r}_t
&=\left[
\mathbf{z}_t^{\pi},
\mathbf{z}_t^v,
\tilde{\mathbf{z}}_t^v,
\mathbf{d}_t,
\bar{\mathbf{p}}_t,
c_t^{\mathrm{map}}
\right],
\label{eq:gru_input}
\end{align}
where $\bar{\mathbf{p}}_t \in [0,1]^2$ is the reported pose
$\hat{\mathbf{p}}_t$ normalized componentwise by the domain
dimensions.

A single-layer GRU ($D_h=96$) feeds a decoder consisting of an $8\times4\times4$ fully connected projection and two
transposed-convolution layers ($8\!\rightarrow16\!\rightarrow2$), producing $\hat{\boldsymbol\mu}_t\in\mathbb R^{2\times21\times21}$. Auxiliary heads predict $q_t$, $\kappa_t$, relative pose, and reconstructed sensing.

The predicted patch is splatted at the reported pose. Its effective reliability and cellwise write mass are \begin{align} \tilde{\boldsymbol{\mu}}_t &= \Pi(\hat{\boldsymbol{\mu}}_t;\hat{\mathbf{p}}_t), \label{eq:splat}\\ \kappa_t^{\mathrm{eff}} &= (1-c_t^{\mathrm{map}}) +c_t^{\mathrm{map}}\kappa_t, \label{eq:kappa_eff}\\ w_t(x,y) &= m_t(x,y)\,q_t\,\kappa_t^{\mathrm{eff}}, \label{eq:fusion_weight} \end{align} where $m_t$ denotes patch support. If $c_t^{\mathrm{map}}\leq\tau_\Psi$, then
$\kappa_t^{\mathrm{eff}}\geq1-\tau_\Psi=0.7$, preserving map
initialization; otherwise, $\kappa_t$ controls the update.

Velocity and evidence are updated pointwise: \begin{align} \boldsymbol{\Omega}_t &= \frac{ \Psi_{t-1}\boldsymbol{\Omega}_{t-1} +w_t\tilde{\boldsymbol{\mu}}_t }{ \Psi_{t-1}+w_t+\epsilon }, \label{eq:omega_update}\\ \Psi_t &= \mathrm{clip}_{[0,1]} \left(\Psi_{t-1}+w_t\right). \label{eq:psi_update} \end{align} Evidence saturation limits visit-count dominance while retaining the ability of later reliable observations to modify the map.

\subsection{Training} \label{sec:training}

We set the patch size to $P=21$ and the map-reference threshold to
$\tau_{\Psi}=0.3$. During training, the simulator provides
observations at both the true and drifted poses. Their alignment
error defines the reliability target
\begin{equation}
e_t=
\left\|
\hat{\mathbf{p}}_t-\mathbf{p}_t^\star
\right\|_2,
\qquad
\kappa_t^{\mathrm{align}}
=
\exp(-e_t/\sigma_{\kappa}),
\quad
\sigma_{\kappa}=5.0~\mathrm{px}.
\label{eq:kappa_align}
\end{equation}

Training proceeds in two stages. In Stage 1, the encoders,
shared GRU, and all prediction heads are trained jointly using
\begin{equation}
\mathcal{L}_{\mathrm{S1}}
=
\lambda_{\mathrm{rec}}\mathcal{L}_{\mathrm{rec}}
+
\lambda_{\Delta p}\mathcal{L}_{\Delta p}
+
\lambda_{\kappa}\mathcal{L}_{\kappa}
+
\lambda_{\mathrm{patch}}\mathcal{L}_{\mathrm{patch}}
+
\lambda_q\mathcal{L}_q .
\label{eq:stage1_loss}
\end{equation}

The terms supervise sensor reconstruction, relative pose, write reliability, velocity-patch prediction, and
informativeness, respectively.

The informativeness target combines support and local flow
structure:
\begin{equation}
q_t^\star
=
0.3\,q_{\mathrm{sup},t}
+
0.7\,q_{\mathrm{struct},t},
\label{eq:q_target}
\end{equation}
where $q_{\mathrm{sup},t}$ is the valid-support fraction of the
target patch and $q_{\mathrm{struct},t}\in[0,1]$ is obtained from
the normalized vector-patch gradient magnitude. The weights were selected via a sweep on the validation split.

In Stage 2, the encoders, patch-decoder weights, and
sensor-reconstruction head are frozen; the GRU and pose,
reliability, and quality heads remain trainable. Because changes
to the recurrent state can still alter decoder outputs, we retain
$\mathcal L_{\mathrm{patch}}$ to prevent reliability calibration
from drifting the patch representation.

\begin{figure*}[!t]
\centering
\includegraphics[width=0.98\textwidth]{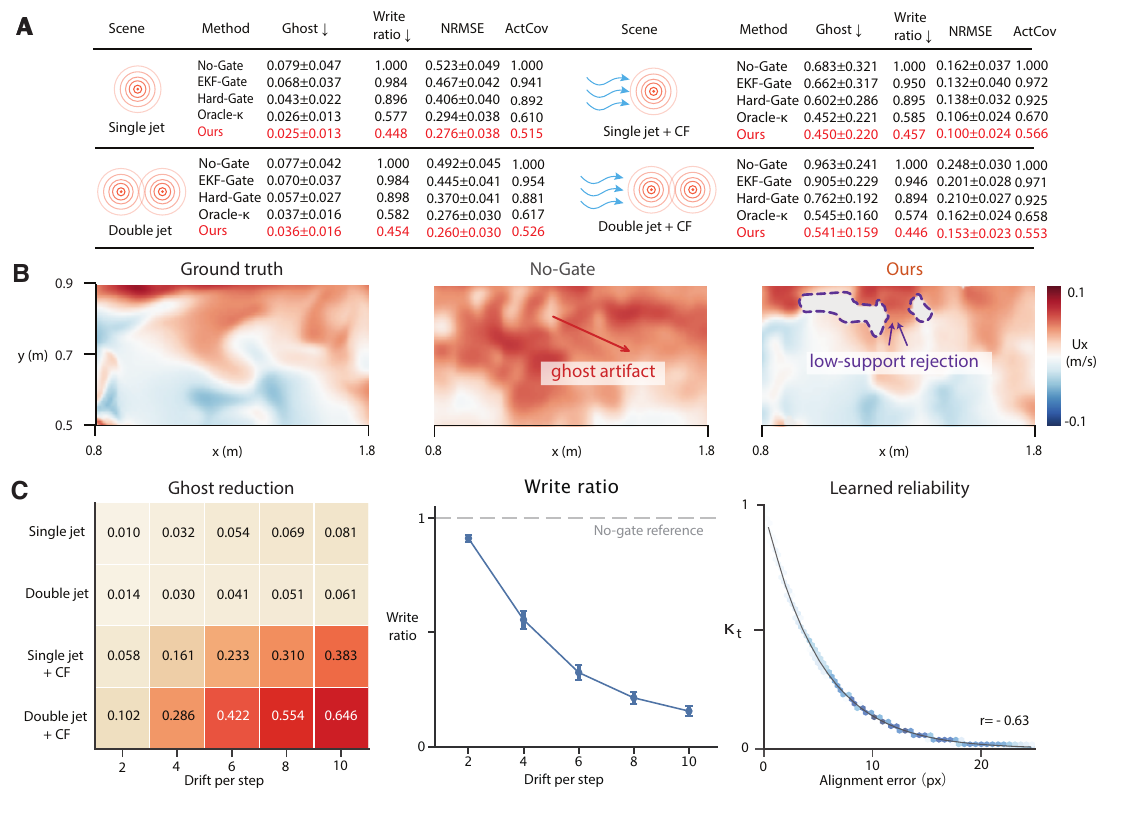}
\vspace{-5pt}

\caption{
Main mapping results.
\textbf{(A)} Ghost, WR, NRMSE, and ActCov at $d=6$ px/step
over 20 held-out scenes per family and three drift seeds.
\textbf{(B)} Representative reconstruction under pose
misregistration; purple contours denote insufficient map support.
\textbf{(C)} Ghost reduction and WR versus drift, and
$\kappa_t$ versus realized alignment error ($r=-0.63$).
}

\label{fig:main_results}
\vspace{-10pt}
\end{figure*}
\section{Experiments} \label{sec:experiments}

We evaluate whether conservative fusion suppresses drift-induced ghost corruption across synthetic flow scenes, whether $\kappa_t$ tracks alignment risk, and whether the learned behavior transfers zero-shot to real onboard measurements.

\subsection{Experimental Setup} \label{sec:exp_setup}

\paragraph{Flow scenes}
We evaluate four $3\,\mathrm{m}\times1\,\mathrm{m}$ FluidX3D
scenes: single jet, double jet, and their crossflow (CF)
variants. Crossflow introduces background advection and greater
spatial ambiguity. The model reconstructs the two-component
velocity field $(u_x,u_y)$. Ghost and NRMSE are reported on
$u_x$, which carries the dominant jet and crossflow structure and is
therefore the component most affected by misregistration; ActCov uses
the velocity magnitude $|u|=\sqrt{u_x^2+u_y^2}$. Training, validation,
and evaluation use scene-disjoint splits.
Evaluation uses 80 held-out scenes (20 per flow family) and three drift seeds per scene and drift level; metrics are reported as mean$\pm$standard deviation across seeds after scene averaging.

\paragraph{Evaluation protocol} All methods receive the same physical trajectory and sensor stream. Each episode follows a predefined lawnmower scan over $\mathcal{D}$. The true sensing pose $\mathbf{p}_t^\star$ is therefore identical across methods. Only the reported pose $\hat{\mathbf{p}}_t$ is corrupted.

This protocol isolates the fusion rule. It prevents differences in navigation policy or sensing coverage from affecting the comparison.

\paragraph{Drift process} Localization drift is cumulative. We perturb the reported motion with a Gaussian random walk: \begin{equation} \begin{aligned} \hat{\mathbf{p}}_0 &= \mathbf{p}^{\star}_0+\boldsymbol{\xi}_0, \qquad \boldsymbol{\xi}_0 \sim \mathcal{N} \left( \mathbf{0}, \sigma_0^2\mathbf{I} \right), \\ \hat{\mathbf{p}}_t &= \mathrm{clip}_{\mathcal{D}} \left[ \hat{\mathbf{p}}_{t-1} + \left( \mathbf{p}^{\star}_t - \mathbf{p}^{\star}_{t-1} \right) + \boldsymbol{\delta}_t \right], \\ \boldsymbol{\delta}_t &\sim \mathcal{N} \left( \mathbf{0}, d^2\mathbf{I} \right). \end{aligned} \label{eq:drift_model} \end{equation} Here, $d$ controls the nominal drift increment in pixels per step. The operator $\mathrm{clip}_{\mathcal{D}}(\cdot)$ keeps the reported write pose inside the domain. We use $\sigma_0=4.0$ px.

Because the accumulated error is stochastic, we also report the realized alignment error $e_t$ defined in Eq.~\eqref{eq:kappa_align}. The main comparison uses $d=6$ px/step, and the sweep uses
$d\in\{2,4,6,8,10\}$ px/step.

\paragraph{Compared methods}
The baselines isolate four forms of write control: no gating,
per-cell innovation filtering, binary gating, and privileged
drift-based gating. We compare:

\begin{itemize}\setlength\itemsep{1pt}

\item \textbf{No-Gate}: the same patch predictor with reliability
gating disabled, $\kappa_t^{\mathrm{eff}}\equiv 1$.

\item \textbf{EKF-Gate}: per-cell extended Kalman filtering with
innovation gating.

\item \textbf{Hard-Gate}: binary write control using
$\kappa_t^{\mathrm{eff}}>\tau$, with $\tau=0.5$.

\item \textbf{Oracle-$\kappa$}: a privileged reference gate based on
the ground-truth alignment error,
\begin{equation}
\kappa_t^{\mathrm{oracle}}
=
\exp\left(
-\frac{
\left\|\hat{\mathbf{p}}_t-\mathbf{p}_t^\star\right\|_2
}{
\sigma_{\kappa,\mathrm{oracle}}
}
\right).
\end{equation}
We set $\sigma_{\kappa,\mathrm{oracle}}=5.0~\mathrm{px}$, matching
the decay scale used for the alignment-reliability target during
training. This baseline is unavailable during deployment and tests
whether ground-truth pose error alone is sufficient to determine
write safety.

\end{itemize}

\paragraph{Metrics}
For each method $m$, we report Ghost, supported-region NRMSE,
active-flow coverage (ActCov), and write ratio (WR).

Let
$\mathcal{Q}=\{(x,y):|u_x^\star(x,y)|
\leq \mathrm{P}_{15}(|u_x^\star|)\}$
be the lowest-flow 15\% of the evaluation grid. The normalized
Ghost score is
\begin{equation}
\mathrm{Ghost}_m
=
\frac{
|\mathcal{Q}|^{-1}
\sum_{(x,y)\in\mathcal{Q}}
|\Omega_{T,x}^{m}(x,y)|
}{
\operatorname{std}(u_x^\star)+\epsilon
}.
\label{eq:ghost_metric}
\end{equation}
This normalization makes Ghost dimensionless.

For the final supported region
$\mathcal{S}_m=\{(x,y):\Psi_T^m(x,y)\geq\tau_\Psi\}$,
we compute
\begin{equation}
\mathrm{NRMSE}_m
=
\frac{
\sqrt{
|\mathcal{S}_m|^{-1}
\sum_{(x,y)\in\mathcal{S}_m}
\left(
\Omega_{T,x}^{m}(x,y)-u_x^\star(x,y)
\right)^2
}
}{
u_{\mathrm{ref}}+\epsilon
},
\label{eq:nrmse_support}
\end{equation}
where
$u_{\mathrm{ref}}=\mathrm{P}_{90}(|u_x^\star|)$
is computed over the evaluation grid and shared across methods.

Using the active-flow region
$\mathcal{A}=\{(x,y):
\|\mathbf{u}^\star(x,y)\|_2\geq0.05~\mathrm{m/s}\}$,
we define
\begin{equation}
\mathrm{ActCov}_m
=
\frac{
|\mathcal{A}\cap\mathcal{S}_{\mathrm{NoGate}}
\cap\mathcal{S}_m|
}{
|\mathcal{A}\cap\mathcal{S}_{\mathrm{NoGate}}|
}.
\label{eq:active_coverage}
\end{equation}
Finally,
\begin{equation}
\mathrm{WR}_m
=
\frac{
\sum_t\sum_{(x,y)}w_t^m(x,y)
}{
\sum_t\sum_{(x,y)}w_t^{\mathrm{NoGate}}(x,y)
}.
\label{eq:write_ratio}
\end{equation}
Lower Ghost and NRMSE indicate cleaner and more accurate maps,
higher ActCov indicates greater active-flow retention, and lower
WR indicates more conservative writing. By definition,
$\mathrm{ActCov}_{\mathrm{NoGate}}
=\mathrm{WR}_{\mathrm{NoGate}}=1$.
For the ablation study, each variant's cellwise write mass is multiplied
by a global constant selected by bisection so that its write ratio
matches that of the full model to within $0.005$.

\subsection{Main Results}
\label{sec:main_results}

Figure~\ref{fig:main_results} summarizes the results at
$d=6$ px per step. Across the four scene families, Ours reduces
the average Ghost score from $0.451$ to $0.263$ ($42\%$) and
NRMSE from $0.356$ to $0.197$, with an average ActCov of $0.540$
and WR of $0.451$. The largest absolute Ghost reductions occur
in the crossflow scenes, where background advection increases
spatial ambiguity and the accumulation of misregistered writes.

EKF-Gate retains nearly all active coverage and write mass
(ActCov $0.960$, WR $0.966$) but provides limited Ghost
reduction. Hard-Gate preserves more coverage than Ours but has
higher average Ghost and NRMSE values of $0.366$ and $0.281$.
The privileged Oracle-$\kappa$ baseline performs comparably
(Ghost $0.265$, NRMSE $0.210$) but writes more
(WR $0.579$ versus $0.451$), indicating that the learned gate
recovers comparable write-safety behavior from
deployment-available inputs at a more conservative operating
point. Figure~\ref{fig:main_results}(B) further shows that Ours
preserves the dominant flow structure by leaving uncertain
regions unsupported, whereas No-Gate accumulates a visible
ghost artifact.

\begin{figure*}[!t]
\centering
\includegraphics[width=0.98\textwidth]{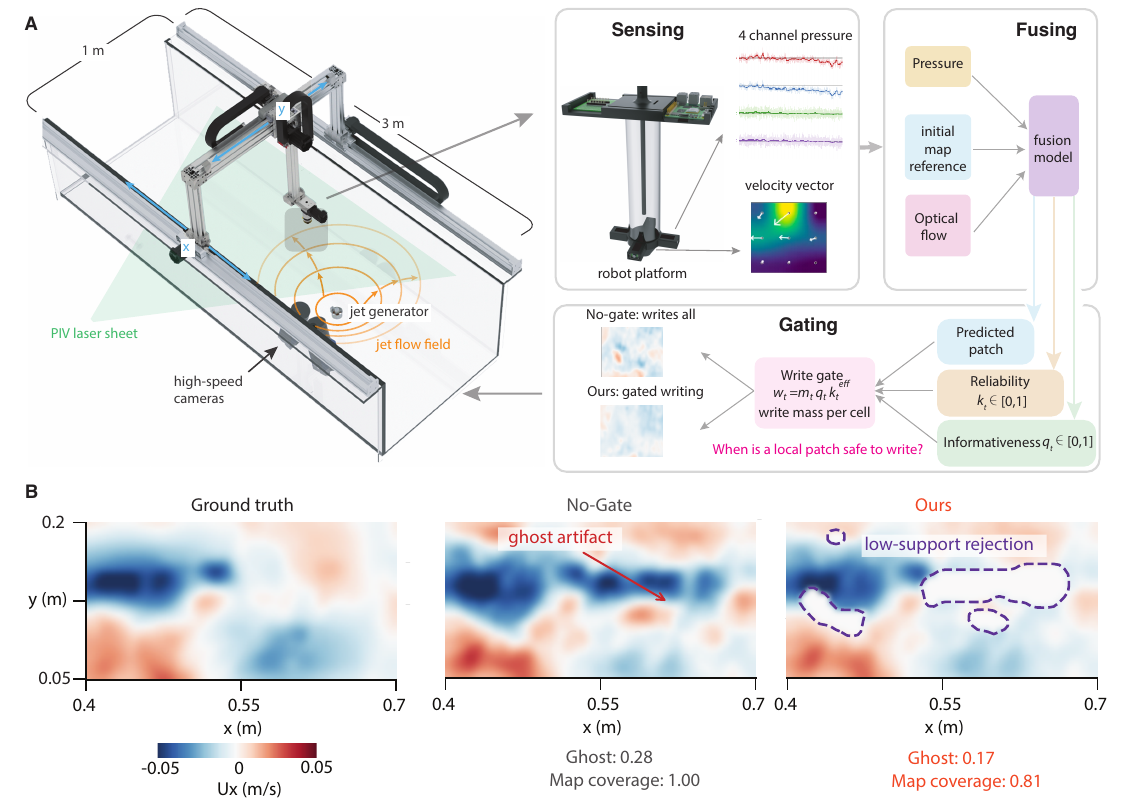}

\caption{
Hardware replay validation.
\textbf{(A)} Water-tank setup and sensing--fusion pipeline;
planar PIV provides the reference field.
\textbf{(B)} Representative reconstruction under injected drift.
No-Gate forms a ghost artifact, whereas gated fusion leaves
uncertain regions unsupported. Displayed maps show one seed;
reported values average three seeds.
}

\label{fig:hardware_validation}
\vspace{-10pt}
\end{figure*}

\begin{table}[t]
\centering
\caption{Component ablation at $d=6$ px/step. Each variant's
write mass is scaled by a single global factor to match the write
ratio of the full model, so that differences reflect where write
mass is placed rather than how much is written.}
\label{tab:ablation}
\small
\begin{tabular}{lcccc}
\toprule
Variant & Ghost $\downarrow$ & WR $\downarrow$ & NRMSE $\downarrow$ & ActCov $\uparrow$ \\
\midrule
Ours (full) & \textbf{0.263} & 0.451 & \textbf{0.197} & \textbf{0.540} \\
No map conditioning & 0.304 & 0.455 & 0.232 & 0.515 \\
No informativeness gate & 0.291 & 0.448 & 0.218 & 0.502 \\
No recurrent state & 0.337 & 0.452 & 0.256 & 0.487 \\
\bottomrule
\end{tabular}
\end{table}

\subsection{Component Ablation}

At matched WR, the full model achieves the lowest Ghost and
NRMSE and the highest ActCov (Table~\ref{tab:ablation}).
Removing the recurrent state causes the largest degradation,
confirming the importance of temporal alignment evidence.
Removing map conditioning primarily worsens Ghost and NRMSE,
whereas removing the informativeness gate causes a larger loss
in active-flow coverage, consistent with their complementary
roles in write safety and write allocation.

\subsection{Mechanism Analysis}
\label{sec:mechanism}

Figure~\ref{fig:main_results}C analyzes how the proposed fusion rule responds to increasing localization drift. The absolute Ghost reduction relative to No-Gate increases with the injected drift level across all
four flow families, with the largest gains observed in the single-jet and double-jet crossflow scenes. These environments contain stronger background advection and greater spatial ambiguity, making them more susceptible to the accumulation of misregistered write-ins. The increasing Ghost reduction is accompanied by a progressively lower write ratio. This indicates that the proposed method becomes more
conservative as alignment risk increases, committing less cumulative
write mass to the evolving map. The improved robustness at larger drift
levels should therefore be interpreted as stronger suppression of unsafe
updates rather than improved reconstruction fidelity under more severe
localization error.

The right panel of Fig.~\ref{fig:main_results}C further shows that the
learned reliability score $\kappa_t$ decreases as the realized
ground-truth alignment error increases, with a Pearson correlation of
$r=-0.63$. Because the true alignment error is used only to construct
training targets and is unavailable during deployment, this result shows
that the learned score preserves the intended alignment-risk ordering
on held-out episodes using only deployment-available inputs. The score
therefore responds systematically to alignment risk rather than acting
as a constant sparsification factor.

\subsection{Robot-Platform Validation}
\label{sec:hardware}

\paragraph{Setup} The hardware experiment tests whether the learned
write-safety behavior transfers to onboard sensing in a realistic flow
environment. We conduct the experiment in a
$3\,\mathrm{m}\times1\,\mathrm{m}$ water tank, where a submerged U5
underwater thruster generates a jet-dominated turbulent flow. The
sensing module is mounted on a linear translation stage and follows a
predefined lawnmower path in a cross-stream plane approximately
$0.3\,\mathrm{m}$ downstream of the thruster.

The module carries four piezoresistive pressure sensors at the arm tips
and a central OV9281 global-shutter camera, which serves as a local
vector-velocity sensor by returning a planar stencil rather than a
single-point measurement. It estimates in-plane velocity from neutrally
buoyant seeding particles using sparse optical flow. A Raspberry Pi 5
records the measurements. The 105.8k-parameter model requires
$12.4\pm2.1$ ms per prediction--fusion step on this platform (500-step
average), dominated by framework overhead rather than arithmetic cost at
this model size. This is negligible compared with the stop-and-acquire
interval of the linear-stage scan, so fusion completes before the next
sensing pose. Two-dimensional PIV on the same plane provides the
reference velocity field, time-averaged over the scan duration.

The real thruster wake is more unsteady and less spatially repeatable
than the synthetic jet flows used for training, though the characteristic
velocity scale is comparable. We therefore deploy the simulation-trained
model zero-shot, with all weights frozen and no fine-tuning or domain
adaptation.

\paragraph{Drift injection} We replay the same physical scan under
controlled localization drift, applying the stochastic perturbation
process from Eq.~\eqref{eq:drift_model} to the recorded stage positions
while the physical trajectory and sensor stream remain unchanged. As in
simulation, this isolates the fusion rule: every method receives
identical measurements and sensing coverage. We evaluate both a
zero-drift reference and an injected-drift condition, the latter
repeated using three independent drift seeds.

\paragraph{Results} The learned write-safety behavior transfers to real sensing. Absolute reconstruction quality is lower than in simulation because of wake unsteadiness, sensor nonlinearities, and PIV--prediction calibration mismatch. The relevant comparison, however, is whether the fusion rule still suppresses drift-induced corruption.

Map coverage is defined as the fraction of cells in the evaluation ROI whose final evidence satisfies \begin{equation} \Psi_T(x,y)\geq\tau_{\Psi}, \qquad \tau_{\Psi}=0.3. \end{equation}Averaged over three independent drift seeds, Ours reduces the
Ghost score from $0.28$ to $0.17$ under injected drift. This
corresponds to a $39\%$ reduction relative to No-Gate. The
reduction is obtained while retaining $81\%$ map coverage,
compared with full coverage for No-Gate. The missing coverage
reflects the intended behavior of conservative fusion: uncertain
regions are left partially unwritten rather than being filled with
spatially misregistered estimates.

Figure~\ref{fig:hardware_validation} shows a representative comparison
within the evaluation ROI: No-Gate accumulates misregistered
measurements into a visible ghost artifact, whereas Ours attenuates
these updates and preserves the dominant wake structure.

Under zero drift, both methods qualitatively recover the principal
structure observed in the PIV reference. Reliability gating does not
visibly suppress the dominant wake when alignment is accurate.
The hardware replay therefore confirms the same trade-off observed
in simulation: conservative writing reduces ghost corruption while
sacrificing a limited amount of map coverage.

\section{Conclusion} \label{sec:conclusion}
A locally plausible flow estimate is not necessarily safe to write into a global map. Under observation aliasing and localization drift, repeated misregistered updates can create persistent ghost structures and corrupt subsequent local-to-global comparisons. We addressed this failure mode with a map-reference-aware conservative fusion framework. The learned write-safety score $\kappa_t$ attenuates uncertain updates while still allowing map initialization when no reliable counterpart is available. The method therefore preserves map integrity where persistent references for loop closure are unavailable.

Across synthetic jet and crossflow environments, the proposed method
reduced average Ghost contamination by $42\%$ relative to No-Gate, and a
zero-shot water-tank replay showed that the same behavior transfers to
real pressure and optical-flow measurements while retaining $81\%$ map
coverage. These results support a broader view of online environmental
mapping: when alignment is uncertain, deciding \emph{whether to write}
can be as important as deciding \emph{what to predict}.


\end{document}